\documentclass[runningheads]{llncs}
\usepackage{graphicx}
\usepackage{amsmath,amssymb} % define this before the line numbering.
\usepackage{color}
\usepackage[width=122mm,left=12mm,paperwidth=146mm,height=193mm,top=12mm,paperheight=217mm]{geometry}
\usepackage{wrapfig}
\usepackage{subcaption}
\usepackage{cite}
\usepackage[colorlinks,citecolor={green}]{hyperref}

\newcommand{\eg}{\emph{e.g. }}
\newcommand{\ie}{\emph{i.e. }}

\begin{document}
% \linenumbers
\pagestyle{headings}
\mainmatter

\title{Revisiting RCNN: On Awakening the Classification Power of Faster RCNN} 
% Replace with your title

\titlerunning{Revisiting RCNN}
% Replace with a meaningful short version of your title

\authorrunning{B. Cheng, Y. Wei, H. Shi, R. Feris, J. Xiong and T. Huang}
% Replace with shorter version of the author list. If there are more authors than fits a line, please use A. Author et al.

\author{Bowen Cheng$^{1}$ \and Yunchao Wei$^{1}\thanks{corresponding author}$  \and Honghui Shi$^{2}$ \and \\
Rogerio Feris$^{2}$ \and Jinjun Xiong$^{2}$ \and Thomas Huang$^{1}$}

%Please write out author names in full in the paper, i.e. full given and family names. 
%If any authors have names that can be parsed into FirstName LastName in multiple ways, please include the correct parsing, in a comment to the volume editors:
%\index{Lastnames, Firstnames}
%(Do not uncomment it, because you may introduce extra index items if you do that, we will use scripts for introducing index entries...)

\institute{
% {\small $^{1}$IFP Group, Beckman Institute at UIUC, IL, USA}\\
{\small $^{1}$University of Illinois at Urbana-Champaign, IL, USA}\\
	\email{ \{bcheng9, yunchao, t-huang1\}@illinois.edu} \\
{\small $^{2}$IBM T.J. Watson Research Center, NY, USA}\\
	\email{Honghui.Shi@ibm.com \{rsferis, jinjun\}@us.ibm.com}
}

\maketitle
\begin{abstract}
Recent region-based object detectors are usually built with separate classification and localization branches on top of shared feature extraction networks. In this paper, we analyze failure cases of state-of-the-art detectors and observe that most \textit{hard false positives} result from classification instead of localization. We conjecture that: 
(1) Shared feature representation is not optimal due to the mismatched goals of feature learning for classification and localization; 
(2) multi-task learning helps, yet optimization of the multi-task loss may result in sub-optimal for individual tasks; 
(3) large receptive field for different scales leads to redundant context information for small objects.
We demonstrate the potential of detector classification power by a simple, effective, and widely-applicable \textit{Decoupled Classification Refinement} (DCR) network. 
DCR samples hard false positives from the base classifier in Faster RCNN and trains a RCNN-styled strong classifier. Experiments show new state-of-the-art results on PASCAL VOC and COCO without any bells and whistles. 

\keywords{Object Detection}
\end{abstract}

\section{Introduction}

Region-based approaches with convolutional neural networks (CNNs) \cite{girshick2014rich,girshick2015fast,ren2015faster,cai2017cascade,xu2017deep,li2017attentive,li2018multistage,li2017perceptual,liang2015towards,wei2018ts2c} have achieved great success in object detection. Such detectors are usually built with separate classification and localization branches on top of shared feature extraction networks, and trained with multi-task loss. In particular, Faster RCNN \cite{ren2015faster} learns one of the first end-to-end two-stage detector with remarkable efficiency and accuracy. Many follow-up works, such as R-FCN \cite{dai2016r}, Feature Pyramid Networks (FPN) \cite{lin2017feature}, Deformable ConvNets (DCN) \cite{dai2017deformable}, have been leading popular detection benchmark in PASCAL VOC \cite{Everingham10} and COCO \cite{lin2014microsoft} datasets in terms of accuracy. Yet, few work has been proposed to study what is the full potential of the classification power in Faster RCNN styled detectors.

\begin{figure}[t]
	\centering
	\includegraphics[width=0.8\textwidth]{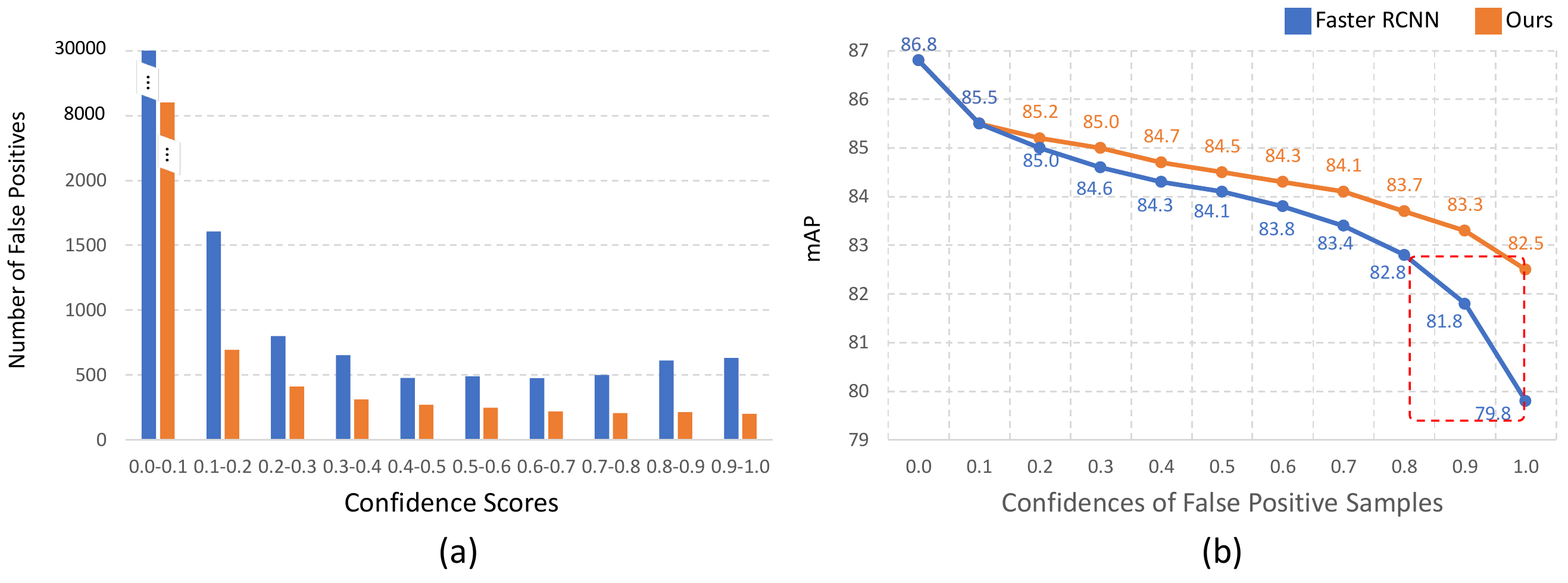}
	\caption{(a) Comparison of the number of false positives in different ranges. (b) Comparison of the mAP gains by progressively removing false positives; from right to left, the detector is performing better as false positives are removed according to their confidence scores.}
	\label{fig:motivation}
%     \vspace{-6mm}
\end{figure}

To answer this question, in this paper, we begin with investigating the key factors affecting the performance of Faster RCNN. As shown in Figure~\ref{fig:motivation} (a), we conduct object detection on PASCAL VOC 2007 using Faster RCNN and count the number of false positive detections in different confidence score intervals (blue). Although only a small percentage of all false positives are predicted with high confidence scores, these samples lead to a significant performance drop in mean average precision (mAP). In particular, we perform an analysis of potential gains in mAP using Faster RCNN: As illustrated in Figure~\ref{fig:motivation} (b), given the detection results from Faster RCNN and a confidence score threshold, we assume that all false positives with predicted confidence score above that threshold were classified correctly and we report the correspondent hypothesized mAP. It is evident that by correcting all false positives, Faster RCNN could, hypothetically, have achieved $86.8\%$ in mAP instead of $79.8\%$. Moreover, even if we only eliminate false positives with high confidences, as indicated in the red box, we can still improve the detection performance significantly by $3.0\%$ mAP, which is a desired yet hard-to-obtain boost for modern object detection systems. 

The above observation motivates our work to alleviate the burden of false positives and improve the classification power of Faster RCNN based detectors.
By scrutinizing the false positives produced by Faster RCNN, we conjecture that such errors are mainly due to three reasons: (1) Shared feature representation for both classification and localization may not be optimal for region proposal classification, the mismatched goals in feature learning lead to the reduced classification power of Faster RCNN; (2) Multi-task learning in general helps to improve the performance of object detectors as shown in Fast RCNN \cite{girshick2015fast} and Faster RCNN, but the joint optimization also leads to  possible sub-optimal to balance the goals of multiple tasks and could not directly utilize the full potential on individual tasks; (3) Receptive fields in deep CNNs such as ResNet-101 \cite{he2016deep} are large, the whole image are usually fully covered for any given region proposals. Such large receptive fields could lead to inferior classification capacity by introducing redundant context information for small objects.

Following the above argument, we propose a simple yet effective approach, named Decoupled Classification Refinement (DCR), to eliminate high-scored false positives and improve the region proposal classification results. DCR decouples the classification and localization tasks in Faster RCNN styled detectors. It takes input from a base classifier, \eg the Faster RCNN, and refine the classification results using a RCNN-styled network. DCR samples \textit{hard false positives}, namely the false positives with high confidence scores, from the base classifier, and then trains a stronger correctional classifier for the classification refinement. Designedly, we do not share any parameters between the Faster RCNN and our DCR module, so that the DCR module can not only utilize the multi-task learning improved results from region proposal networks (RPN) and bounding box regression tasks, but also better optimize the newly introduced module to address the challenging classification cases.

\begin{figure}[t]
	\centering
	\includegraphics[width=0.8\textwidth]{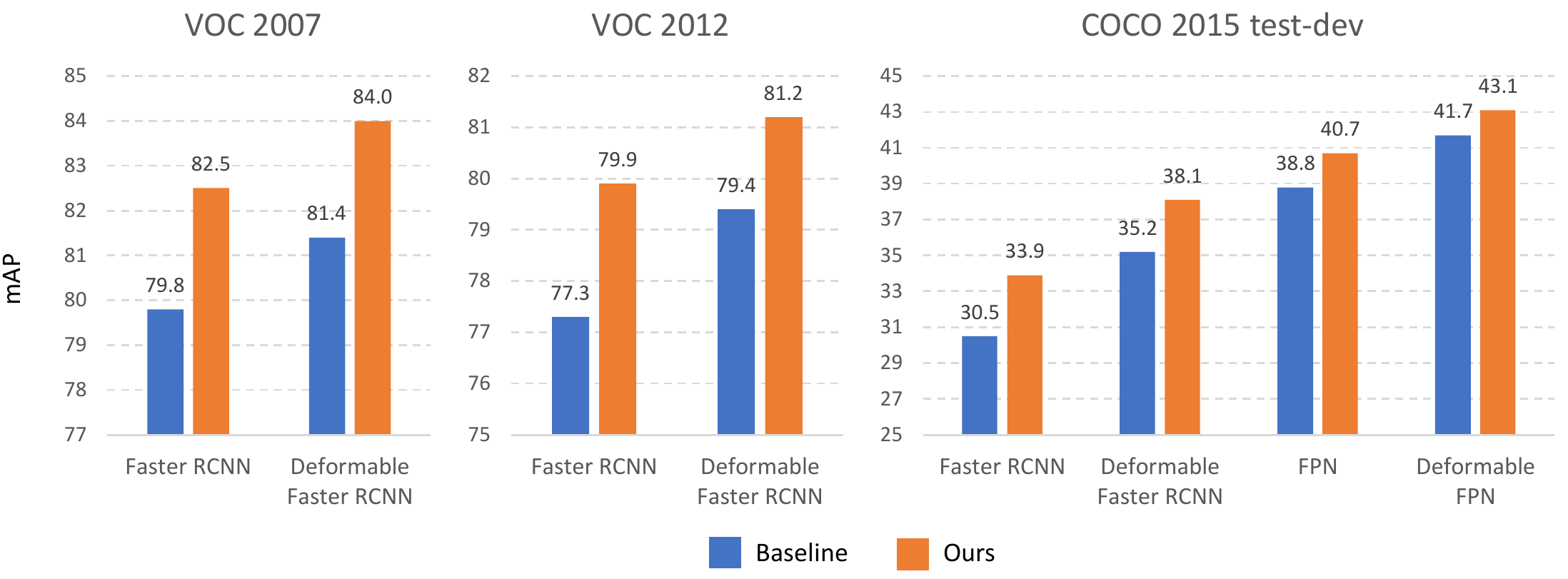}
	\caption{Comparison of our approach and baseline in terms of different Faster RCNN series and benchmarks.}
	\label{fig:improvement}
%     \vspace{-6mm}
\end{figure}

We conduct extensive experiments based on different Faster RCNN styled detectors (\ie Faster RCNN, Deformable ConvNets, FPN) and benchmarks (\ie PASCAL VOC 2007 \& 2012, COCO) to demonstrate the effectiveness of our proposed simple solution in enhancing the detection performance by alleviating hard false positives. 
As shown in Figure~\ref{fig:motivation} (a), our approach can significantly reduce the number of hard false positives and boost the detection performance by $2.7\%$ in mAP on PASCAL VOC 2007 over a strong baseline as indicated in Figure~\ref{fig:motivation} (b).
All of our experiment results demonstrate that our proposed DCR module can provide consistent improvements over various detection baselines, as shown in Figure~\ref{fig:improvement}. Our contributions are threefold:
\begin{enumerate}
\item
We analyze the error modes of region-based object detectors and formulate the hypotheses that might cause these failure cases.
\item
We propose a set of design principles to improve the classification power of Faster RCNN styled object detectors along with the DCR module based on the proposed design principles.
\item
Our DCR module consistently brings significant performance boost to strong object detectors on popular benchmarks. In particular, following common practice (ResNet-101 as backbone), we achieve mAP of $84.0\%$ and $81.2\%$ on the classic PASCAL VOC 2007 and 2012, respectively, and $43.1\%$ on the more challenging COCO2015 \emph{test-dev}, which are the new state-of-the-art.
%We show how to apply the DCR module to refine object detector and observed consistent improvement over several challenge benchmarks. We achieve (percent) the new state-of-the-art result on Pascal VOC 2007 dataset using only VOC training images and achieve a consistently (percent) improvement on COCO dataset.
\end{enumerate}
% \vspace{-6mm}

\section{Related Work}
% \vspace{-2mm}
\subsubsection{Object Detection}
Recent CNN based object detectors can generally be categorized into two-stage and single stage. One of the first two-stage detector is RCNN \cite{girshick2014rich}, where selective search \cite{uijlings2013selective} is used to generate a set of region proposals for object candidates, then a deep neural network to extract feature vector of each region followed by SVM classifiers. SPPNet \cite{he2014spatial} improves the efficiency of RCNN by sharing feature extraction stage and use spatial pyramid pooling to extract fixed length feature for each proposal. Fast RCNN \cite{girshick2015fast} improves over SPPNet by introducing an differentiable ROI Pooling operation to train the network end-to-end. Faster RCNN \cite{ren2015faster} embeds the region proposal step into a Region Proposal Network (RPN) that further reduce the proposal generation time. R-FCN \cite{dai2016r} proposed a position sensitive ROI Pooling (PSROI Pooling) that can share computation among classification branch and bounding box regression branch. Deformable ConvNets (DCN) \cite{dai2017deformable} further add deformable convolutions and deformable ROI Pooling operations, that use learned offsets to adjust position of each sampling bin in naive convolutions and ROI Pooling, to Faster RCNN. Feature Pyramid Networks (FPN) \cite{lin2017feature} add a top-down path with lateral connections to build a pyramid of features with different resolutions and attach detection heads to each level of the feature pyramid for making prediction. Finer feature maps are more useful for detecting small objects and thus a significant boost in small object detection is observed with FPN. Most of the current state-of-the-art object detectors are two-stage detectors based of Faster RCNN, because two-stage object detectors produce more accurate results and are easier to optimize. However, two-stage detectors are slow in speed and require very large input sizes due to the ROI Pooling operation. Aimed at achieving real time object detectors, one-stage method, such as OverFeat \cite{sermanet2013overfeat}, SSD \cite{liu2016ssd,fu2017dssd} and YOLO \cite{redmon2016you,redmon2017yolo9000}, predict object classes and locations directly. Though single stage methods are much faster than two-stage methods, their results are inferior and they need more extra data and extensive data augmentation to get better results. Our paper follows the method of two-stage detectors \cite{girshick2014rich,girshick2015fast,ren2015faster}, but with a main focus on analyzing reasons why detectors make mistakes.
% \vspace{-6mm}
\subsubsection{Classifier Cascade}
The method of classifier cascade commonly trains a stage classifier using misclassified examples from a previous classifier. This has been used a lot for object detection in the past. The Viola Jones Algorithm \cite{viola2004robust} for face detection used a hard cascades by Adaboost \cite{freund1997decision}, where a strong region classifier is built with cascade of many weak classifier focusing attentions on different features and if any of the weak classifier rejects the window, there will be no more process. Soft cascades \cite{bourdev2005robust} improved \cite{viola2004robust} built each weak classifier based on the output of all previous classifiers. Deformable Part Model (DPM) \cite{felzenszwalb2010object} used a cascade of parts method where a root filter on coarse feature covering the entire object is combined with some part filters on fine feature with greater localization accuracy. More recently, Li et al. \cite{li2015convolutional} proposed the Convolutional Neural Network Cascade for fast face detection. Our paper proposed a method similar to the classifier cascade idea, however, they are different in the following aspects. The classifier cascade aims at producing an efficient classifier (mainly in speed) by cascade weak but fast classifiers and the weak classifiers are used to reject examples. In comparison, our method aims at improving the overall system accuracy, where exactly two strong classifiers are cascaded and they work together to make more accurate predictions. More recently, Cascade RCNN \cite{cai2017cascade} proposes training object detector in a cascade manner with gradually increased IoU threshold to assign ground truth labels to align the testing metric, ie. average mAP with IOU 0.5:0.05:0.95.

%  DCN producing better results by stronger localization powers.
%, where a CNN classifier with very small input size of $12\times12$ is first used to reject 90\% of the detection windows followed with bounding box regression with $12\times12$ boxes as input, a cascaded CNN classifier with input size $24\times24$ is used to further reject 90\% of the remaining windows followed with another bounding box regression network with $24\times24$ input

\begin{figure}[t]
	\centering
	\includegraphics[width=0.8\textwidth]{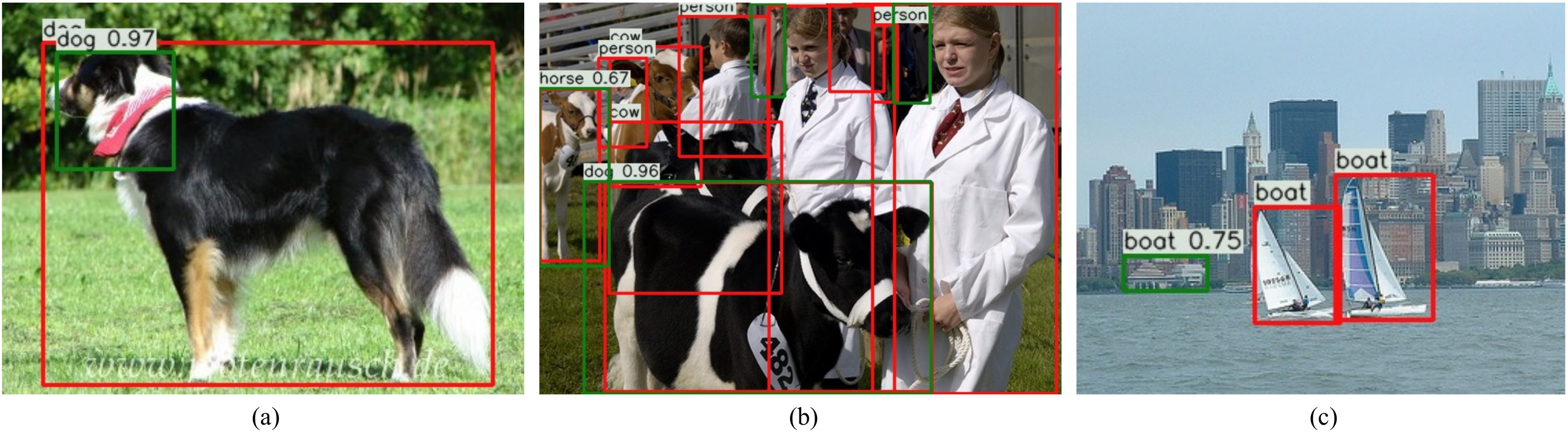}
	\caption{Demonstration of hard false positives. Results are generate by Faster RCNN with 2 fully connected layer (2fc) as detector head \cite{ren2015faster,lin2017feature}, red boxes are ground truth, green boxes are hard false positives with scores higher than 0.3; (a) boxes covering only part of objects with high confidences; (b) incorrect classification due to similar objects; (c) misclassified backgrounds.}
	\label{fig:FP}
%     \vspace{-6mm}
\end{figure}
\section{Problems with Faster RCNN}

Faster RCNN produces 3 typical types of hard false positives, as shown in Fig \ref{fig:FP}:
(1) The classification is correct but the overlap between the predicted box and ground truth has low IoU, \eg $<0.5$ in Fig \ref{fig:FP} (a). This type of false negative boxes usually cover the most discriminative part  and have enough information to predict the correct classes due to translation invariance. 
(2) Incorrect classification for predicted boxes but the IoU with ground truth are large enough , \eg in Fig \ref{fig:FP} (b). It happens mainly because some classes share similar discriminative parts and the predicted box does not align well with the true object and happens to cover only the discriminative parts of confusion. Another reason is that the classifier used in the detector is not strong enough to distinguish between two similar classes. (3) the detection is a ``confident" background, meaning that there is no intersection or small intersection with ground truth box but classifier's confidence score is large, \eg in Fig \ref{fig:FP} (c). Most of the background pattern in this case is similar to its predicted class and the classifier is too weak to distinguish. Another reason for this case is that the receptive field is fixed and it is too large for some box that it covers the actual object in its receptive field. In Fig \ref{fig:FP} (c), the misclassified background is close to a ground truth box (the left boat), and the large receptive field (covers more than 1000 pixels in ResNet-101) might ``sees'' too much object features to make the wrong prediction. Given above analysis, we can conclude that the hard false positives are mainly caused by the suboptimal classifier embedded in the detector. The reasons may be that: (1) feature sharing between classification and localization, (2) optimizing the sum of classification loss and localization loss, and (3) detector's receptive field does not change according to the size of objects.

%\subsection{Potential of Faster RCNN}
%We study the potential of Faster RCNN by evaluate its mAP gains in the ideal case. Suppose that we have an ideal classifier that rejects an false positive with confidence larger than a threshold. A threshold of 1.0 means the classifier does not reject any false positives and a threshold of 0.0 means the classifier rejects all false positives. We study the threshold decreasing from 1.0 to 0.0 with a step of 0.1 and results is shown in Fig \ref{fig:motivation} (c). The horizontal axis is the threshold of the ideal classifier and the vertical axis is the resulting mAP. The baseline of Faster RCNN is 79.8\% and if all false positives are rejected, the maximum performance of Faster RCNN is achieved with mAP 86.8\%. Notice that mAP increase significantly from 1.0 to 0.8 (3.0\%) as well as from 0.1 to 0.0 (1.3\%). The former gain comes from the fact that false positives with large influence even if the number is small and the latter gain comes from the fact that there are too many false positives.

%\subsection{Problems with Detector}
%From the visualization, we find that the classifier in a detector is not well optimized, as it cannot distinguish between two similar parts of objects. We hypothesize that hard false positives are caused by the following three reasons: (1) feature sharing between classification and localization, (2) optimizing the sum of classification loss and localization loss, and (3) detector's receptive field does not change according to the size of objects.

% \vspace{-3mm}
\subsubsection{Problem with Feature Sharing}
\label{problem1}
Detector backbones are usually adapted from image classification model and pre-trained on large image classification dataset. These backbones are original designed to learn scale invariant features for classification. Scale invariance is achieved by adding sub-sampling layers, \eg max pooling, and data augmentation, \eg random crop. Detectors place a classification branch and localization branch on top of the same backbone, however, classification needs \textbf{translation invariant} feature whereas localization needs \textbf{translation covariant} feature. During fine-tuning, the localization branch will force the backbone to gradually learn translation covariant feature, which might potentially down-grade the performance of classifier.
%\bowen{which may potentially be a problem.} \notsure{During fine-tuning, a small learning rate is used because we do not want the feature extractor to change much, but the localization branch will force the backbone feature extractor to gradually learn translation covariant feature which might down-grade the performance of classifier.}
% \vspace{-3mm}
\subsubsection{Problem with Optimization}
\label{problem2}
%As discussed in Section \ref{pipeline}, current object detection framework optimizes the sum of a classification loss and a localization loss denoted as $L_{detection}=L_{cls}+L_{bbox}$ in a Multi-Task Learning manner. 

Faster RCNN series are built with a feature extractor as backbone and two task-specified branches for classifying regions and the other for localizing correct locations. Denote loss functions for classification and localization as $L_{cls}$ and $L_{bbox}$, respectively. Then, the optimization of Faster RCNN series is to address a Multi-Task Learning (MTL) problem by minimizing the sum of two loss functions: $L_{detection}=L_{cls}+L_{bbox}$. However, the optimization might converge to a compromising suboptimal of two tasks by simultaneously considering the sum of two losses, instead of each of them. 
% \bowen{However, the role of MTL in training object detection framework has hardly been studied and it is only discussed in few literals, \eg \cite{girshick2015fast}.}
% \delete{At the early stage of training, both losses will drop simultaneously. As the training progresses, the two losses will compete against each other and the drop of one loss might force to increase the other loss; however, the sum of the two losses will still decrease. The contest might end up with converging to a suboptimal of two tasks.}

Originally, such a MTL manner is found to be effective and observed improvement over state-wise learning in Fast(er) RCNN works. However, MTL for object detection is not studied under the recent powerful classification backbones, \eg ResNets. Concretely, we hypothesize that MTL may work well based on a weak backbone (\eg AlexNet or VGG16). As the backbone is getting stronger, the powerful classification capacity within the backbone may not be fully exploited and MTL becomes the bottleneck.

%Originally, MTL is used to jointly train multiple classification task to boost the performance of each individual task. MTL is also found to be effective when applied to unrelated tasks but there is still a lack of theory. \bowen{We argue that the uncertainty about MTL is problem in current object detection frameworks as we are not sure whether the current training pipeline is optimal.} \delete{MTL is first applied to object detection and observed improvement over stage-wise learning in Fast RCNN paper, but it is not studied under recent detection frameworks. We hypothesize that MTL works for Fast RCNN because the backbone and detection head is very weak (\ie VGG16 is used). As backbone and detection head is getting stronger, \eg ResNet-101, the capacity is not fully utilized and MTL becomes the bottleneck.}
% \vspace{-3mm}
\subsubsection{Problem with Receptive Field}
\label{problem3}
Deep convolutional neural networks have fixed receptive fields. For image classification, inputs are usually cropped and resized to have fixed sizes, \eg $224 \times 224$, and network is designed to have a receptive field little larger than the input region. However, since contexts are cropped and objects with different scales are resized, the ``effective receptive field'' is covering the whole object.

Unlike image classification task where a single large object is in the center of a image, objects in detection task have various sizes over arbitrary locations. In Faster RCNN, the ROI pooling is introduced to crop object from 2-D convolutional feature maps to a 1-D fixed size representation for the following classification, which results in fixed receptive field (\ie the network is attending to a fixed-size window of the input image). In such a case, objects have various sizes and the fixed receptive field will introduce different amount of context. For a small object, the context might be too large for the network to focus on the object whereas for a large object, the receptive field might be too small that the network is looking at part of the object. Although some works introduce multi-scale features by aggregating features with different receptive field, the number of sizes is still too small comparing with the number various sizes of objects.

%%bowen
%Unlike image classification task where a single large object is in the center of a image, objects in detection task have various sizes, cover only small parts in a image and distribute not necessarily the center part of a image. Classification is done one the cropped features. In a two-stage detector, an ROI Pooling operation is introduced to crop object features to a fixed size representation for further processing. In a one-stage detector, each column feature on the feature map is usually responsible for objects with centers in its cell. The problem is that the receptive field is fixed, that is, the network is attending to a fixed-size window of the input image. Objects have various sizes and the fixed receptive field will introduce different amount of context. For a small object, the context might be too large for the network to focus on the object whereas for a large object, the receptive field might be too small that the network is looking at part of the object. 

\section{Revisiting RCNN for Improving Faster RCNN}
% \vspace{-2mm}
\label{DCR}
% In this section, we first introduce three principals for designing new object detectors that are able to suppress hard false positives. Then we provide a simple ye effective solution by augmenting a decoupled classification refinement module to the current state-of-the-art object detectors.
% \wyc{[RCNN is an outdated detector. Why it is a good choice for fitting the design principals?]}

%In this section, we look back closely into the RCNN \cite{girshick2014rich} method, and give an in depth analysis of both its disadvantages (why RCNN does not work well?) and advantages (some good design principals). We find the idea of RCNN can be used as a ``complement'' to improve Faster RCNN. Based on this finding, we solve problems of Faster RCNN by providing a simple yet effective decoupled classification refinement module, that can be easily added to any current state-of-the-art object detectors.

% \begin{wrapfigure} {l} {0.5\textwidth}
% % 	\vspace{-6mm}
% 	\includegraphics[width=0.5\textwidth]{fig/framework2.pdf}
% 	\caption{Left: base detector (\eg Faster RCNN). Right: our proposed Decoupled Classification Refinement (DCR) module.}
% 	\label{fig:framework}
% % 	\vspace{-6mm}
% \end{wrapfigure}

\begin{figure}[t]
% 	\vspace{-6mm}
	\centering
	\includegraphics[width=0.6\textwidth]{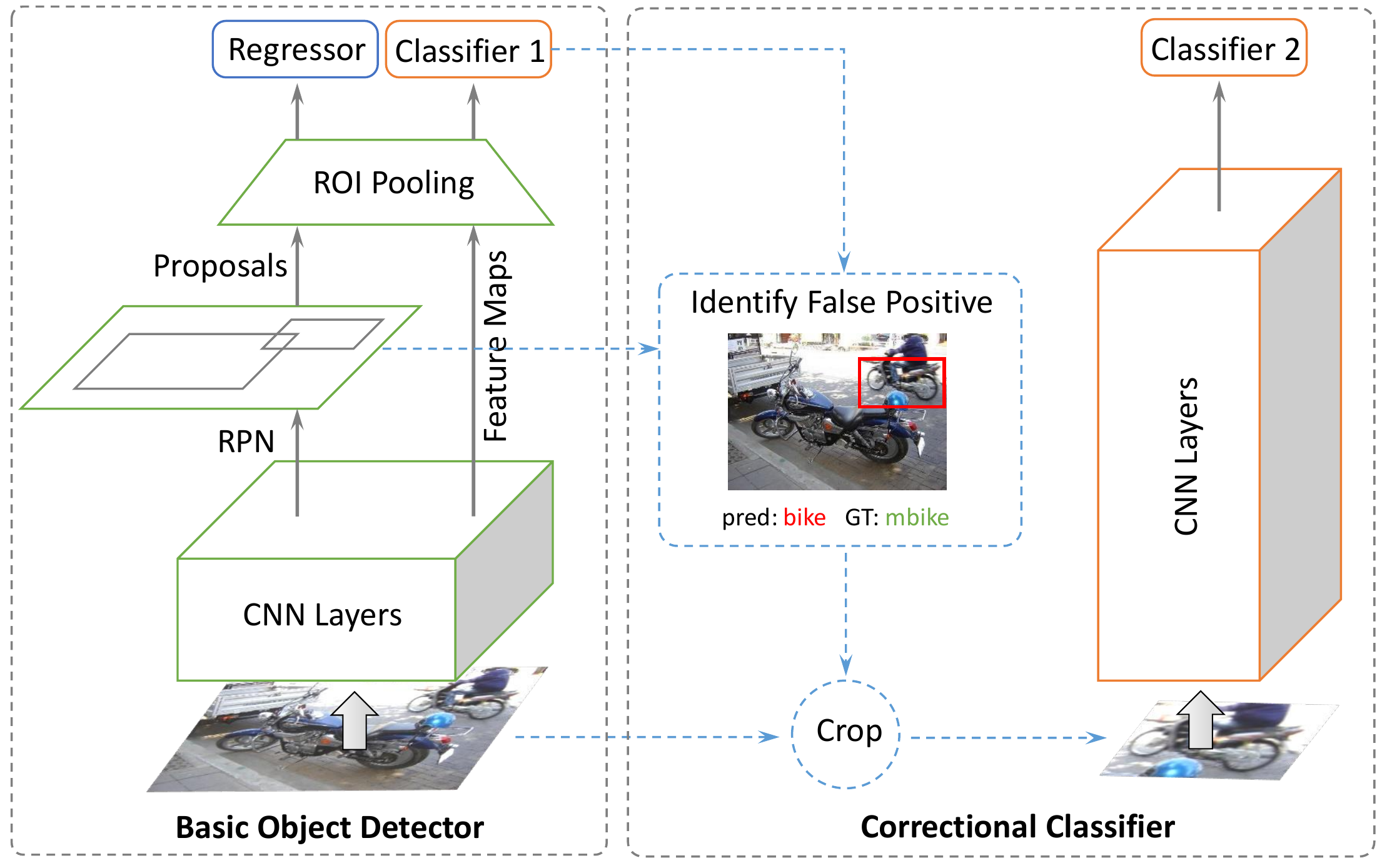}
	\caption{Left: base detector (\eg Faster RCNN). Right: our proposed Decoupled Classification Refinement (DCR) module.}
	\label{fig:framework}
% 	\vspace{-6mm}
\end{figure}

In this section, we look back closely into the classic RCNN \cite{girshick2014rich} method, and give an in-depth analysis of why RCNN can be used as a ``complement'' to improve Faster RCNN. Based on our findings, we provide a simple yet effective decoupled classification refinement module, that can be easily added to any current state-of-the-art object detectors to provide performance improvements.
% \vspace{-3mm}
\subsection{Learning from RCNN Design}

We train a modified RCNN with ResNet-50 as backbone and Faster RCNN predictions as region proposals. We find that with RCNN along, the detection result is deteriorated by more than 30\% (from 79.8\% to 44.7\%)! Since RCNN does not modify box coordinate, the inferior result means worse classification. We find that many boxes having small intersections with an object are classified as that object instead of the background which Faster RCNN predicts. Based on this finding, we hypothesize that the drawback of RCNN is mainly root from that classification model is pre-trained without awaring object location. Since ResNet-50 is trained on ImageNet in multi-crop manner, no matter how much the intersection of the crop to the object is, classifier is encouraged to predict that class. This leads to the classifier in RCNN being ``too strong'' for proposal classification, and this is why RCNN needs a carefully tuned sampling strategy, \ie a ratio of 1:3 of fg to bg. Straightforwardly, we are interested whether RCNN is ``strong'' enough to correct hard negatives. We make a minor modification to multiply RCNN classification score with Faster RCNN classification score and observe a boost of 1.9\% (from 79.8\% to 81.7\%)! Thus, we consider that RCNN can be seen as a compliment of Faster RCNN in the following sense: the classifier of Faster RCNN is weaker but aware of object location whereas the classifier of RCNN is unaware of object location but stronger. Based on our findings, we propose the following three principals to design a better object detector.

% \subsection{Design Principals}
% \vspace{-3mm}
\subsubsection{Decoupled Features}
%RCNN decouples the feature used for object proposal (selective search \cite{uijlings2013selective} is used) and classification. 
%\notsure{In Section \ref{problem1}, we hypothesize that sharing the same feature between classification task and localization task will result in learning a feature representation that is sub-optimal for either task.} 
Current detectors still place classification head and localization head on the same backbone, hence we propose that classification head and localization head should not share parameter (as the analysis given in Section \ref{problem1}), resulted in a decoupled feature using pattern by RCNN.
% \vspace{-3mm}
\subsubsection{Decoupled Optimization}
RCNN also decouples the optimization for object proposal and classification. %\delete{Object detectors are trained with MTL to minimize the sum of two losses, in general. We conjecture in Section \ref{problem2} that the two tasks might be unrelated, \ie classification and localization have nothing in common, and MTL might end up with sub-optimal solution for both tasks.} 
{In this paper, we make a small change in optimization. We propose a novel two-stage training where, instead of optimizing the sum of classification and localization loss, we optimize the concatenation of classification and localization loss, $L_{detection} = [L_{cls} + L_{bbox}, L_{cls}]$, where each entry is being optimized independently in two steps.}

\subsubsection{Adaptive Receptive Field}
The most important advantage of RCNN is that its receptive field always covers the whole ROI, 
\ie the receptive field size adjusts according to the size of the object by cropping and resizing each proposal to a fixed size. We agree that context information may be important for precise detection, however, we conjuncture that different amount of context introduced by fixed receptive filed might cause different performance to different sizes of objects. It leads to our last proposed principal that a detector should an adaptive receptive field that can change according to the size of objects it attends to. In this principal, the context introduced for each object should be proportional to its size, but how to decide the amount of context still remains an open question to be studied in the future. Another advantage of adaptive receptive field is that its features are well aligned to objects. Current detectors make predictions at hight-level, coarse feature maps usually having a large stride, \eg a stride of 16 or 32 is used in Faster RCNN, due to sub-sampling operations. The sub-sampling introduces unaligned features, \eg one cell shift on a feature map of stride 32 leads to 32 pixels shift on the image, and defects the predictions. With adaptive receptive field, the detector always attends to the entire object resulting in an aligned feature to make predictions. RCNN gives us a simple way to achieve adaptive receptive field, but how to find a more efficient way to do so remains an interesting problem needs studying.

% \begin{figure}[t]
% 	\centering
% 	\includegraphics[width=0.7\textwidth]{fig/framework2.pdf}
% 	\caption{Left: base detector (\eg Faster RCNN). Right: our proposed Decoupled Classification Refinement (DCR) module.}
% 	\label{fig:framework}
%     \vspace{-5mm}
% \end{figure}

\subsection{Decoupled Classification Refinement (DCR)}
%Following these principals, we propose a DCR module that can be easily augmented to Faster RCNN as well as any object detector to build a stronger detector. The overall pipeline is shown in Fig \ref{fig:framework}. On the left part of Fig \ref{fig:framework} is the original Faster RCNN, it consists of backbone feature extractor, a region proposal network and a detection head with a classification branch and a bounding box regression branch on top of each ROI feature. On the right part of Fig \ref{fig:framework} is our proposed DCR module. It mainly consists a crop and resize layer and a strong classifier. The crop and resize layer takes two input, the original image and boxes produced by Faster RCNN, crops boxes on the original image and feed them to the classifier after resizing them to a predefined size. Region scores of DCR module (Classifier 2) is aggregated with region scores of Faster RCNN (Classifier 1) by element-wise product to form the final score of each region.

Following these principals, we propose a DCR module that can be easily augmented to Faster RCNN as well as any object detector to build a stronger detector. The overall pipeline is shown in Fig \ref{fig:framework}. The left part and the right part are the original Faster RCNN and our proposed DCR module, respectively. In particular, DCR mainly consists a crop-resize layer and a strong classifier. The crop-resize layer takes two inputs, the original image and boxes produced by Faster RCNN, crops boxes on the original image and feeds them to the strong classifier after resizing them to a predefined size. Region scores of DCR module (Classifier 2) is aggregated with region scores of Faster RCNN (Classifier 1) by element-wise product to form the final score of each region. The two parts are trained separately in this paper and the scores are only combined during test time.

The DCR module does not share any feature with the detector backbone in order to preserve the quality of classification-aimed translation invariance feature. Furthermore, there is no error propagation between DCR module and the base detector, thus the optimization of one loss does not affect the other. This in turn results in a decoupled pattern where the base detector is focused more on localization whereas the DCR module focuses more on classification. DCR module introduces adaptive receptive field by resizing boxes to a predefined size. Noticed that this processing is very similar to moving an ROI Pooling from final feature maps to the image, however, it is quite different than doing ROI Pooling on feature maps. Even though the final output feature map sizes are the same, features from ROI Pooling sees larger region because objects embedded in an image has richer context. We truncated the context by cropping objects directly on the image and the network cannot see context outside object regions.

%Although the receptive field is fixed, we cropped box regions and resize then to align objects with different sizes to the fixed receptive field. 
% \vspace{-3mm}
\subsection{Training}
\label{training}

% Since there is no error propagates from the DCR module to Faster RCNN, we train our object detector in a two-step manner. First, we train Faster RCNN to converge. Then, we train our DCR module on mini-batches sampled from hard false positives of Faster RCNN.

% DCR module is first pre-trained on ImageNet dataset \cite{deng2009imagenet}, after pre-training, the last fully connected layer is re-initialized with normal distribution (zero mean and std 0.01) and output $C+1$ class scores ($C$ is the number of object classes and we append one more class to for background). We follow the image-centric method \cite{girshick2015fast} to sample $N$ images with a total mini-batch size of $R$ boxes, \ie $R/N$ boxes per image. We use $N=1$ and $R=32$ throughout experiments. We use a different sampling heuristic that we sample not only foreground and background boxes but also hard false positive \textbf{uniformly}. Because we do not want to apply any prior knowledge to impose unnecessary bias on classifier. However, we observed that boxes from the same image have little variance. Thus, we fix Batch Normalization layer with ImageNet training set statistics. The newly added linear classifier (fully connected layer) is set with 10 times the base learning rate as we want to preserve translation invariance features learned on the ImageNet dataset.

Since there is no error propagates from the DCR module to Faster RCNN, we train our object detector in a two-step manner. First, we train Faster RCNN to converge. Then, we train our DCR module on mini-batches sampled from hard false positives of Faster RCNN. Parameters of DCR module are pre-trained by ImageNet dataset \cite{deng2009imagenet}. We follow the image-centric method \cite{girshick2015fast} to sample $N$ images with a total mini-batch size of $R$ boxes, \ie $R/N$ boxes per image. We use $N=1$ and $R=32$ throughout experiments. We use a different sampling heuristic that we sample not only foreground and background boxes but also hard false positive \textbf{uniformly}. Because we do not want to apply any prior knowledge to impose unnecessary bias on classifier. However, we observed that boxes from the same image have little variance. Thus, we fix Batch Normalization layer with ImageNet training set statistics. The newly added linear classifier (fully connected layer) is set with 10 times of the base learning rate since we want to preserve translation invariance features learned on the ImageNet dataset.
\section{Experiments}
% \vspace{-3mm}
\subsection{Implementation Details}
We train base detectors, \eg Faster RCNN, following their original implementations. We use default settings in \ref{training} for DCR module, we use ROI size $224 \times 224$ and use a threshold of 0.3 to identify hard false positives. Our DCR module is first pre-trained on ILSVRC 2012 \cite{deng2009imagenet}. In fine-tuning, we set the initial learning rate to $0.0001$ \emph{w.r.t}. one GPU and weight decay of $0.0001$. We follow linear scaling rule in \cite{goyal2017accurate} for data parallelism on multiple GPUs and use 4 GPUs for PASCAL VOC and 8 GPUs for COCO. Synchronized SGD with momentum $0.9$ is used as optimizer. No data augmentation except horizontal flip is used.
% \vspace{-4mm}
\subsection{Ablation Studies on PASCAL VOC}
% \vspace{-2mm}
%%%%%%%%%%%%%%%%%%%%%%%%%%%%%%%%%%%%%
\begin{table}[t]
	\scriptsize
	\centering
	%\resizebox{1\textwidth}{!}{
	\begin{minipage}[t]{0.22\textwidth}
		\vspace{0pt}
		\centering
		\begin{tabular}{l|c}
			Sample method & mAP \\
			\hline
			Baseline & 79.8 \\
			\hline
			Random & 81.8\\ 
			FP Only & 81.4 \\
			FP+FG & 81.6 \\
			FP+BG & 80.3 \\
			FP+FG+BG & \textbf{82.3} \\
			RCNN-like & 81.7 \\	
		\end{tabular}
		%		\caption{(a)}
		%			\caption*{ (a) Ablation study on sampling heuristic.}
	\end{minipage}% <---- don't forget this %
	\begin{minipage}[t]{0.22\textwidth}
		\vspace{0pt}
		\centering	
		\begin{tabular}{l|c}
			FP Score & mAP \\
			\hline
			Baseline & 79.8 \\
			\hline
			0.20 & 82.2\\
			0.25 & 81.9 \\
			0.30 & \textbf{82.3} \\
			0.35 & 82.2 \\
			0.40 & 82.0 \\
		\end{tabular}
		%			\caption*{(b) Ablation study on threshold for hard false positive score}
	\end{minipage}	
	\begin{minipage}[t]{.22\textwidth}
		\vspace{0pt}
		\centering		
		\begin{tabular}{l|c}
			Sample size & mAP \\
			\hline
			Baseline & 79.8 \\
			\hline
			8 Boxes & 82.0 \\
			16 Boxes & 82.1 \\
			32 Boxes & \textbf{82.3} \\
			64 Boxes & 82.1 \\			
		\end{tabular}
		%			\caption*{(c) Ablation study on sampling number.}
	\end{minipage}% <---- don't forget this %
	\begin{minipage}[t]{.3\textwidth}
		\vspace{0pt}
		\begin{tabular}{l|c|c}
			ROI scale & mAP & Test Time \\
			\hline
			Baseline & 79.8 & 0.0855 \\
			\hline
			$56 \times 56$ & 80.6 & 0.0525\\
			$112 \times 112$ & 82.0 & 0.1454\\
			$224 \times 224$ & \textbf{82.3} & 0.5481\\
			$320 \times 320$ & 82.0 & 1.0465\\
		\end{tabular}
		%			\caption{Ablation study on box scale.}			
	\end{minipage}
	%	}	

	\begin{minipage}[b]{.22\textwidth}
		\centering
		(a)
	\end{minipage}
	\begin{minipage}[b]{.22\textwidth}
		\centering
		(b)
	\end{minipage}
	\begin{minipage}[b]{.22\textwidth}
		\centering
		(c)
	\end{minipage}
	\begin{minipage}[b]{.3\textwidth}
		\centering
		(d)
	\end{minipage}
	%	\resizebox{1\textwidth}{!}{		
	\begin{minipage}[t]{.35\textwidth}
		\vspace{0pt}
		\begin{tabular}{l|c|c} 
			DCR Depth & mAP & Test Time \\
			\hline
			Baseline & 79.8 & 0.0855  \\
			\hline
			18 & 81.4 & 0.1941 \\
			34 & 81.9 & 0.3144 \\
			50 & 82.3 & 0.5481 \\
			101 & 82.3 & 0.9570 \\
			152 & \textbf{82.5} & 1.3900 \\
		\end{tabular}
		%		\caption{Ablation study on classifier depth.}		
	\end{minipage}
	\begin{minipage}[t]{.22\textwidth}
		\vspace{0pt}
		\begin{tabular}{l|c}
			Base detector & mAP \\
			\hline
			Faster & 79.8 \\
			Faster+DCR & 82.3\\
			\hline
			DCN & 81.4 \\
			DCN+DCR & 83.2\\
		\end{tabular}
		%		\caption{Ablation study on base detector.}		
	\end{minipage}
	\begin{minipage}[t]{.35\textwidth}		
		\vspace{0pt}
		\begin{tabular}{l|c}
			Model capacity & mAP \\
			\hline
			Faster w/ Res101 & 79.8 \\
			Faster w/ Res152 & 80.3 \\
            Faster Ensemble & 81.1 \\
			Faster w/ Res101+DCR-50 & 82.3\\
		\end{tabular}
		%		\caption{Ablation study on model capacity.}
	\end{minipage}	
	
	\begin{minipage}[b]{.35\textwidth}
		\centering
		(e)
	\end{minipage}
	\begin{minipage}[b]{.22\textwidth}
		\centering
		(f)
	\end{minipage}
	\begin{minipage}[b]{.35\textwidth}
		\centering
		(g)
	\end{minipage}
	%	}
	
	\caption{Ablation studies results. Evaluate on PASCAL VOC2007 test set. Baseline is Faster RCNN with ResNet-101 as backbone. DCR module uses ResNet-50. (a) Ablation study on sampling heuristics. (b) Ablation study on threshold for defining hard false positives. (c) Ablation study on sampling size. (d) Ablation study on ROI scale and test time (measured in seconds/image). (e) Ablation study on depth of DCR module and test time (measured in seconds/image). (f) DCR module with difference base detectors. Faster denotes Faster RCNN and DCN denotes Deformable Faster RCNN, both use ResNet-101 as backbone. (g) Comparison of Faster RCNN with same size as Faster RCNN + DCR.}
	\label{ablation}
%     \vspace{-8mm}
\end{table}

We comprehensively evaluate our method on the PASCAL VOC detection benchmark \cite{Everingham10}. We use the union of VOC 2007 trainval and VOC 2012 trainval as well as their horizontal flip as training data and evaluate results on the VOC 2007 test set. We primarily evaluate the detection mAP with IoU 0.5 (mAP@0.5). Unless otherwise stated, all ablation studies are performed with 
ResNet-50 as classifier for our DCR module.
% \vspace{-3mm}
\subsubsection{Ablation study on sampling heuristic}
We compare results with different sampling heuristic in training DCR module:
\begin{itemize}
\item 
random sample: a minibatch of ROIs are randomly sampled for each image
\item
hard false positive only: a minibatch of ROIs that are hard postives are sampled for each image
\item
hard false positive and background: a minibatch of ROIs that are either hard postives or background are sampled for each image
\item
hard false positive and foreground: a minibatch of ROIs that are either hard postives or foreground are sampled for each image
\item
hard false positive, background and foreground: the difference with random sample heuristic is that we ignore easy false positives during training.
\item
RCNN-like: we follow the Fast RCNN's sampling heuristic, we sample two images per GPU and 64 ROIs per image with fg:bg=1:3.
\end{itemize}

Results are shown in Table \ref{ablation} (a). We find that the result is insensitive to sampling heuristic. Even with random sampling, an improvement of 2.0\% in mAP is achieved. With only hard false positive, the DCR achieves an improvement of 1.6\% already. Adding foreground examples only further gains a 0.2\% increase. Adding background examples to false negatives harms the performance by a large margin of 1.1\%. We hypothesize that this is because comparing to false positives, background examples dominating in most images results in a classifier bias to predicting background. This finding demonstrate the importance of hard negative in DCR training. Unlike RCNN-like detectors, we do not make any assumption of the distribution of hard false positives, foregrounds and backgrounds. To balance the training of classifier, we simply uniformly sample from the union set of hard false positives, foregrounds and backgrounds. This uniform sample heuristic gives the largest gain of 2.5\% mAP. We also compare our training with RCNN-like training. Training with RCNN-like sampling heuristic with fg:bg=1:3 only gains a margin of 1.9\%.
% \vspace{-3mm}
\subsubsection{Ablation study on other hyperparameters}

We compare results with different threshold for defining hard false positive: [0.2, 0.25, 0.3, 0.35, 0.4]. Results are shown in Table \ref{ablation} (b). We find that the results are quite insensitive to threshold of hard false positives and we argue that this is due to our robust uniform sampling heuristic. With hard false positive threshold of 0.3, the performance is the best with a gain of 2.5\%.
We also compare the influence of size of sampled RoIs during training: [8, 16, 32, 64]. Results are shown in Table \ref{ablation} (c). Surprisingly, the difference of best and worst performance is only 0.3\%, meaning our method is highly insensitive to the sampling size. With smaller sample size, the training is more efficient without severe drop in performance.

% \vspace{-3mm}
\subsubsection{Speed and accuracy trade-off}
There are in general two ways to reduce inference speed, one is to reduce the size of input and the other one is to reduce the depth of the network. We compare 4 input sizes: $56\times56$, $112\times112$, $224\times224$, $320\times320$ as well as 5 depth choices: 18, 34, 50, 101, 152 and their speed. Results are shown in Table \ref{ablation} (d) and (e). The test speed is linearly related to the area of input image size and there is a severe drop in accuracy if the image size is too small, \eg $56\times56$. For the depth of classifier, deeper model results in more accurate predictions but also more test time. We also notice that the accuracy is correlated with the classification accuracy of classification model, which can be used as a guideline for selecting DCR module.
% \vspace{-3mm}
\subsubsection{Generalization to more advanced object detectors}
We evaluate the DCR module on Faster RCNN and advanced Deformable Convolution Nets (DCN) \cite{dai2017deformable}. Results are shown in Table \ref{ablation} (f). Although DCN is already among one of the most accurate detectors, its classifier still produces hard false positives and our proposed DCR module is effective in eliminating those hard false positives.
% \vspace{-3mm}
\subsubsection{Where is the gain coming from?}
One interesting question is where the accuracy gain comes from. 
Since we add a large convolutional network on top of the object detector, does the gain simply comes from more parameters? Or, is DCR an ensemble of two detectors?
To answer this question, we compare the results of Faster RCNN with ResNet-152 as backbone (denoted Faster-152) and Faster RCNN with ResNet-101 backbone + DCR-50 (denoted Faster-101+DCR-50) and results are shown in Table \ref{ablation} (g). Since the DCR module is simply a classifier, the two network have approximately the same number of parameters. However, we only observe a marginal gain of 0.5\% with Faster-152 while our Faster-101+DCR-50 has a much larger gain of 2.5\%. To show DCR is not simply then ensemble to two Faster RCNNs, we further ensemble Faster RCNN with ResNet-101 and ResNet-152 and the result is 81.1\% which is still 1.1\% worse than our Faster-101+DCR-50 model. This means that the capacity does not merely come from more parameters or ensemble of two detectors.

% \vspace{-3mm}
\subsection{PASCAL VOC Results}
\begin{table*}[tb]\setlength{\tabcolsep}{1pt}
	\centering
	%\tiny
    \resizebox{1\textwidth}{!}{
\begin{tabular}{l|c|cccccccccccccccccccc}
\tiny Method  & mAP & \rotatebox{90}{aero} & \rotatebox{90}{bike} & \rotatebox{90}{bird} & \rotatebox{90}{boat} & \rotatebox{90}{bottle} & \rotatebox{90}{bus} & \rotatebox{90}{car} & \rotatebox{90}{cat} & \rotatebox{90}{chair} & \rotatebox{90}{cow} & \rotatebox{90}{table} & \rotatebox{90}{dog} & \rotatebox{90}{horse} & \rotatebox{90}{mbike} & \rotatebox{90}{person} & \rotatebox{90}{plant} & \rotatebox{90}{sheep} & \rotatebox{90}{sofa} & \rotatebox{90}{train} & \rotatebox{90}{tv} \\
        \hline

      Faster~\cite{he2016deep} 
      & 76.4 & 79.8 & 80.7 & 76.2 & 68.3 & 55.9 & 85.1 & 85.3 & 89.8 & 56.7 & 87.8 & 69.4 & 88.3 & 88.9 & 80.9 & 78.4 &  41.7 & 78.6 & 79.8 & 85.3 & 72.0\\
     
     R-FCN~\cite{dai2016r} & 80.5 & 79.9 & 87.2 & 81.5 & 72.0 & 69.8 & 86.8 & 88.5 & \textbf{89.8} & 67.0 & 88.1 &  74.5 & 89.8 & \textbf{90.6} & 79.9 & 81.2 & 53.7 & 81.8 & 81.5 & 85.9 & 79.9\\
     
     SSD~\cite{liu2016ssd,fu2017dssd} & 80.6 & 84.3 & 87.6 & \textbf{82.6} & 71.6 & 59.0 & 88.2 & 88.1 & 89.3 & 64.4 & 85.6 & 76.2 & 88.5 & 88.9 & \textbf{87.5} & 83.0 & 53.6 & 83.9 & 82.2 & 87.2 & 81.3 \\
     
      DSSD~\cite{fu2017dssd} 
      & 81.5 & 86.6 & 86.2 & \textbf{82.6} & 74.9 & 62.5 &89.0 & 88.7 & 88.8 & 65.2 & 87.0 & \textbf{78.7} & 88.2 & 89.0 & \textbf{87.5} & 83.7 & 51.1 & 86.3 & 81.6 & 85.7 & 83.7 \\

\hline
     Faster (2fc)& 79.8 & 79.6 & 87.5 & 79.5 & 72.8 & 66.7 & 88.5 & 88.0 & 88.9 & 64.5 & 84.8 & 71.9 & 88.7 & 88.2 & 84.8 & 79.8 & 53.8 & 80.3 & 81.4 & \textbf{87.9} & 78.5\\
     Faster-Ours (2fc) & 82.5 & 80.5 & \textbf{89.2} & 80.2 & 75.1 & 74.8 & 79.8 & 89.4 & 89.7 & 70.1 & 88.9 & 76.0 & 89.5 & 89.9 & 86.9 & 80.4 & 57.4 & 86.2 & 83.5 & 87.2 & \textbf{85.3}\\
     
     \hline
     DCN (2fc) & 81.4 & 83.9 & 85.4 & 80.1 & 75.9 & 68.8 & 88.4 & 88.6 & 89.2 & 68.0 & 87.2 & 75.5 & 89.5 & 89.0 & 86.3 & 84.8 & 54.1 & 85.2 & 82.6 & 86.2 & 80.3\\
     DCN-Ours (2fc) & \textbf{84.0} & \textbf{89.3} & 88.7 & 80.5 & \textbf{77.7} & \textbf{76.3} & \textbf{90.1} & \textbf{89.6} & \textbf{89.8} & \textbf{72.9} & \textbf{89.2} & 77.8 & \textbf{90.1} & 90.0 & \textbf{87.5} & \textbf{87.2} & \textbf{58.6} & \textbf{88.2} & \textbf{84.3} & 87.5 & 85.0\\

		\end{tabular}
        }
		\caption{\textbf{PASCAL VOC2007 \texttt{test} detection results.}}
\label{tab:voc07}
% \vspace{-8mm}
\end{table*}

\begin{table*}[htb]\setlength{\tabcolsep}{1pt}
	\centering
	%\tiny
    \resizebox{1\textwidth}{!}{
\begin{tabular}{l|c|cccccccccccccccccccc}
 Method  & mAP & \rotatebox{90}{aero} & \rotatebox{90}{bike} & \rotatebox{90}{bird} & \rotatebox{90}{boat} & \rotatebox{90}{bottle} & \rotatebox{90}{bus} & \rotatebox{90}{car} & \rotatebox{90}{cat} & \rotatebox{90}{chair} & \rotatebox{90}{cow} & \rotatebox{90}{table} & \rotatebox{90}{dog} & \rotatebox{90}{horse} & \rotatebox{90}{mbike} & \rotatebox{90}{person} & \rotatebox{90}{plant} & \rotatebox{90}{sheep} & \rotatebox{90}{sofa} & \rotatebox{90}{train} & \rotatebox{90}{tv} \\

        \hline
        Faster~\cite{he2016deep} & 73.8 & 86.5 & 81.6 & 77.2 & 58.0 & 51.0 & 78.6 & 76.6 & 93.2 & 48.6 & 80.4 & 59.0 & 92.1 & 85.3 & 84.8 & 80.7 & 48.1 & 77.3 & 66.5 & 84.7 & 65.6 \\

       R-FCN~\cite{dai2016r} &  77.6 & 86.9 & 83.4 & 81.5& 63.8& 62.4 & 81.6 & 81.1 & 93.1 & 58.0 & 83.8 & 60.8& 92.7 & 86.0 & 84.6 & 84.4 & 59.0 & 80.8 & 68.6& 86.1 & 72.9 \\

	SSD~\cite{liu2016ssd,fu2017dssd} & 79.4 & 90.7 & \textbf{87.3} & 78.3 & 66.3 & 56.5 & 84.1 & 83.7 &  94.2 & 62.9 & 84.5 & \textbf{66.3} & 92.9 & \textbf{88.6} & 87.9 & 85.7 & 55.1 & 83.6  & \textbf{74.3} & 88.2 & 76.8 \\
   
      DSSD~\cite{fu2017dssd} & 80.0 & \textbf{92.1} & 86.6 & 80.3 & 68.7 & 58.2 & 84.3 & \textbf{85.0} & \textbf{94.6} & 63.3 & 85.9 & 65.6 & 93.0 & 88.5 & 87.8 & 86.4 & 57.4 & 85.2 & 73.4 & 87.8 & 76.8  \\
     
     \hline
     Faster (2fc)& 77.3 & 87.3 & 82.6 & 78.8 & 66.8 & 59.8 & 82.5 & 80.3 & 92.6 & 58.8 & 82.3 & 61.4 & 91.3 & 86.3 & 84.3 & 84.6 & 57.3 & 80.9 & 68.3 & 87.5 & 71.4\\
     Faster-Ours (2fc) & 79.9 & 89.1 & 84.6 & 81.6 & 70.9 & 66.1 & \textbf{84.4} & 83.8 & 93.7 & 61.5 & 85.2 & 63.0 & 92.8 & 87.1 & 86.4 & 86.3 & 62.9 & 84.1 & 69.6 & 87.8 & \textbf{76.9}\\

     \hline
     DCN (2fc) & 79.4 & 87.9 & 86.2 & 81.6 & 71.1 & 62.1 & 83.1 & 83.0 & 94.2 & 61.0 & 84.5 & 63.9 & 93.1 & 87.9 & 87.2 & 86.1 & 60.4 & 84.0 & 70.5 & \textbf{89.0} & 72.1\\
     DCN-Ours (2fc) & \textbf{81.2} & 89.6 & 86.7 & \textbf{83.8} & \textbf{72.8} & \textbf{68.4} & 83.7 & \textbf{85.0} & 94.5 & \textbf{64.1} & \textbf{86.6} & 66.1 & \textbf{94.3} & 88.5 & \textbf{88.5} & \textbf{87.2} & \textbf{63.7} & \textbf{85.6} & 71.4 & 88.1 & 76.1\\

		\end{tabular}
        }
		\caption{\textbf{PASCAL VOC2012 \texttt{test} detection results.} }
\label{tab:voc12}
% \vspace{-8mm}
\end{table*}

%\subsubsection{VOC 2007}
\textbf{VOC 2007} We use a union of VOC2007 trainval and VOC2012 trainval for training and we test on VOC2007 test. We use the default training setting and ResNet-152 as classifier for the DCR module. We train our model for 7 epochs and reduce learning rate by $\frac{1}{10}$ after 4.83 epochs. Results are shown in Table \ref{tab:voc07}. Notice that based on DCN as base detector, our single DCR module achieves the new state-of-the-art result of 84.0\% without using extra data (\eg COCO data), multi scale training/testing, ensemble or other post processing tricks.\\
% \vspace{-3mm}
%\subsubsection{VOC 2012}

\noindent
\textbf{VOC 2012} We use a union of VOC2007 trainvaltest and VOC2012 trainval for training and we test on VOC2012 test. We use the same training setting of VOC2007. Results are shown in Table \ref{tab:voc12}. Our model DCN-DCR is the first to achieve over 81.0\% on the VOC2012 test set. The new state-of-the-art 81.2\% is achieved using only single model, without any post processing tricks.
% \vspace{-3mm}

% \vspace{-2mm}
\subsection{COCO Results}
% \vspace{-1mm}
All experiments on COCO follow the default settings and use ResNet-152 for DCR module. We train our model for 8 epochs on the COCO dataset and reduce the learning rate by $\frac{1}{10}$ after 5.33 epochs. We report results on two different partition of COCO dataset. One partition is training on the union set of COCO2014 train and COCO2014 val35k together with 115k images and evaluate results on the COCO2014 minival with 5k images held out from the COCO2014 val. The other partition is training on the standard COCO2014 trainval with 120k images and evaluate on the COCO2015 test-dev. We use Faster RCNN \cite{ren2015faster}, Feature Pyramid Networks (FPN) \cite{lin2017feature} and the Deformable ConvNets \cite{dai2017deformable} as base detectors. 
\begin{table*}[tb]\setlength{\tabcolsep}{1pt}
	\centering
	\tiny
    \resizebox{1\textwidth}{!}{
\begin{tabular}{l|l|ccc|ccc}
 Method & Backbone & AP & $\text{AP}_{50}$ & $\text{AP}_{75}$ & $\text{AP}_{\text{S}}$ & $\text{AP}_{\text{M}}$ & $\text{AP}_{\text{L}}$ \\
        \hline
        
     Faster (2fc) & ResNet-101& 30.0 & 50.9 & 30.9 & 9.9 & 33.0 & 49.1 \\
     Faster-Ours (2fc) & ResNet-101 + ResNet-152 & 33.1 & 56.3 & 34.2 & 13.8 & 36.2 & 51.5 \\
     
     \hline
     DCN (2fc) & ResNet-101 & 34.4 & 53.8 & 37.2 & 14.4 & 37.7 & 53.1 \\
      DCN-Ours (2fc) & ResNet-101 + ResNet-152 & 37.2 & 58.6 & 39.9 & 17.3 & 41.2 & 55.5 \\
      
      \hline
     FPN & ResNet-101 & 38.2 & 61.1 & 41.9 & 21.8 & 42.3 & 50.3 \\
      FPN-Ours & ResNet-101 + ResNet-152 & 40.2 & 63.8 & 44.0 & 24.3 & 43.9 & 52.6 \\
      
      \hline
     FPN-DCN & ResNet-101 & 41.4 & 63.5 & 45.3 & 24.4 & 45.0 & 55.1 \\
      FPN-DCN-Ours & ResNet-101 + ResNet-152 & \textbf{42.6} & \textbf{65.3} & \textbf{46.5} & \textbf{26.4} & \textbf{46.1} & \textbf{56.4} \\

		\end{tabular}
        }
		\caption{\textbf{COCO2014 \texttt{minival} detection results.}}
        % All detectors use ResNet-101 as backbone and DCR modules use ResNet-152 as base classifier.
\label{tab:coco-minival}
% \vspace{-5mm}
\end{table*}
\begin{table*}[htb]\setlength{\tabcolsep}{1pt}
	\centering
	\tiny
    \resizebox{1\textwidth}{!}{
\begin{tabular}{l|l|ccc|ccc}
 Method & Backbone & AP & $\text{AP}_{50}$ & $\text{AP}_{75}$ & $\text{AP}_{\text{S}}$ & $\text{AP}_{\text{M}}$ & $\text{AP}_{\text{L}}$ \\
 \hline
%  YOLOv2~\cite{redmon2016yolo9000} & DarkNet-19 \cite{redmon2016yolo9000}
%   & 21.6 & 44.0 & 19.2 & 5.0 & 22.4 & 35.5 \\
 SSD~\cite{liu2016ssd,fu2017dssd} & ResNet-101-SSD
  & 31.2 & 50.4 & 33.3 & 10.2 & 34.5 & 49.8 \\
 DSSD513~\cite{fu2017dssd} & ResNet-101-DSSD
%   & 33.2 & 53.3 & 35.2 & 13.0 & 35.4 & 51.1 \\
%  Faster+++~\cite{he2016deep} & ResNet-101-C4
%   & 34.9 & 55.7 & 37.4 & 15.6 & 38.7 & 50.9\\
%   G-RMI~\cite{huang2017speed} & Inception-ResNet-v2 \cite{szegedy2017inception}
%   & 34.7 & 55.5 & 36.7 & 13.5 & 38.1 & 52.0\\
%  FPN~\cite{lin2017feature} & ResNet-101-FPN
  & 36.2 & 59.1 & 39.0 & 18.2 & 39.0 & 48.2\\
  Mask RCNN~\cite{he2017mask} & ResNeXt-101-FPN~\cite{xie2017aggregated}
  & 39.8 & 62.3 & 43.4 & 22.1 & 43.2 & 51.2 \\
 RetinaNet~\cite{lin2017focal} & ResNeXt-101-FPN
  & 40.8 & 61.1 & 44.1 & 24.1 & 44.2 & 51.2 \\
 
        \hline
        
     Faster (2fc) & ResNet-101 & 30.5 & 52.2 & 31.8 & 9.7 & 32.3 & 48.3 \\
     Faster-Ours (2fc) & ResNet-101 + ResNet-152 & 33.9 & 57.9 & 35.3 & 14.0 & 36.1 & 50.8 \\
     
     \hline
     DCN (2fc) & ResNet-101 & 35.2 & 55.1 & 38.2 & 14.6 & 37.4 & 52.6 \\
      DCN-Ours (2fc) & ResNet-101 + ResNet-152 & 38.1 & 59.7 & 41.1 & 17.9 & 41.2 & 54.7 \\
      
      \hline
     FPN & ResNet-101 & 38.8 & 61.7 & 42.6 & 21.9 & 42.1 & 49.7 \\
      FPN-Ours & ResNet-101 + ResNet-152 & 40.7 & 64.4 & 44.6 & 24.3 & 43.7 & 51.9 \\
      
      \hline
     FPN-DCN & ResNet-101 & 41.7 & 64.0 & 45.9 & 23.7 & 44.7 & 53.4 \\
      FPN-DCN-Ours & ResNet-101 + ResNet-152 & \textbf{43.1} & \textbf{66.1} & \textbf{47.3} & \textbf{25.8} & \textbf{45.9} & \textbf{55.3} \\

		\end{tabular}
        }
		\caption{\textbf{COCO2015 \texttt{test-dev} detection results.}}
        % Comparison with state-of-the-arts reported in recent publications with different backbones.
\label{tab:coco-test}
% \vspace{-8mm}
\end{table*}
% \vspace{-3mm}
%\subsubsection{COCO minival}

\noindent
\textbf{COCO minival} Results are shown in Table \ref{tab:coco-minival}. Our DCR module improves Faster RCNN by 3.1\% from 30.0\% to 33.1\% in COCO AP metric. Faster RCNN with DCN is improved by 2.8\% from 34.4\% to 37.2\% and FPN is improved by 2.0\% from 38.2\% to 40.2\%. Notice that FPN+DCN is the base detector by top-3 teams in the COCO2017 detection challenge, but there is still an improvement of 1.2\% from 41.4\% to 42.6\%. This observation shows that currently there is no perfect detector that does not produce hard false positives.\\

%\subsubsection{COCO test-dev}
\noindent
\textbf{COCO test-dev} Results are shown in Table \ref{tab:coco-test}. The trend is similar to that on the COCO minival, with Faster RCNN improved from 30.5\% to 33.9\%, Faster RCNN+DCN improved from 35.2\% to 38.1\%, FPN improved from 38.8\% to 40.7\% and FPN+DCN improved from 41.7\% to 43.1\%. We also compare our results with recent state-of-the-arts reported in publications and our best model achieves state-of-the-art result on COCO2015 test-dev with ResNet as backbone.

% \vspace{-5mm}
% \subsection{Discussions}

% Our DCR module demonstrates extremely good performance in suppressing false positives. Fig \ref{fig:motivation} (a) compares total number of false positives on the VOC2007 test set. With our DCR module, the number of hard false is reduced by almost three times (orange). 
% %We also use \cite{hoiem2012diagnosing} to analyze detection results, more figures and discussions are in supplementary material. The main issue with the DCR module is the running time. 
% The inference time of DCR module is proportional to the number of proposals, input size and network depth. Table \ref{ablation} (d), (e) compare the running time and the best model (DCR with depth 152) runs slower than the baseline Faster RCNN at the speed of 1.39 s/image on 1080 Ti GPU. However, this paper focuses more on the analysis of failure case of object detectors and accuracy boost, improvement to speed will be studied in the future.

% \vspace{-5mm}
\section{Conclusion}
% \vspace{-2mm}
In this paper, we analyze error modes of state-of-the-art region-based object detectors and study their potentials in accuracy improvement. We hypothesize that good object detectors should be designed following three principles: decoupled features, decoupled optimization and adaptive receptive field. Based on these principles, we propose a simple, effective and widely-applicable DCR module that achieves new state-of-the-art. In the future, we will further study what architecture makes a good object detector, adaptive feature representation in multi-task learning, and efficiency improvement of our DCR module.

% \vspace{3mm}
\noindent
\textbf{Acknowledgements.} This work is in part supported by IBM-ILLINOIS Center for Cognitive Computing Systems Research (C3SR) - a research collaboration as part of the IBM AI Horizons Network; 
and by the Intelligence Advanced Research Projects Activity (IARPA) via Department of Interior/ Interior Business Center (DOI/IBC) contract number D17PC00341. The U.S. Government is authorized to reproduce and distribute reprints for Governmental purposes notwithstanding any copyright annotation thereon.  Disclaimer: The views and conclusions contained herein are those of the authors and should not be interpreted as necessarily representing the official policies or endorsements, either expressed or implied, of IARPA, DOI/IBC, or the U.S. Government.
%and by IARPA-BAA-16-13 Deep Intermodal Video Analytics (DIVA). 
%and by the Intelligence Advanced Research Projects Activity (IARPA) via Department of Interior/ Interior Business Center (DOI/IBC) contract number D17PC00341.
We thank Jiashi Feng for helpful discussions.

\clearpage

\title{Supplementary Materials for Revisiting RCNN: On Awakening the Classification Power of Faster RCNN} 
% Replace with your title

\titlerunning{Supplementary materials}
% Replace with a meaningful short version of your title

\authorrunning{B. Cheng, Y. Wei, H. Shi, R. Feris, J. Xiong and T. Huang}
% Replace with shorter version of the author list. If there are more authors than fits a line, please use A. Author et al.

\author{Bowen Cheng$^{1}$ \and Yunchao Wei$^{1}\thanks{corresponding author}$  \and Honghui Shi$^{2}$ \and \\
Rogerio Feris$^{2}$ \and Jinjun Xiong$^{2}$ \and Thomas Huang$^{1}$}

%Please write out author names in full in the paper, i.e. full given and family names. 
%If any authors have names that can be parsed into FirstName LastName in multiple ways, please include the correct parsing, in a comment to the volume editors:
%\index{Lastnames, Firstnames}
%(Do not uncomment it, because you may introduce extra index items if you do that, we will use scripts for introducing index entries...)

\institute{
% {\small $^{1}$IFP Group, Beckman Institute at UIUC, IL, USA}\\
{\small $^{1}$University of Illinois at Urbana-Champaign, IL, USA}\\
	\email{ \{bcheng9, yunchao, t-huang1\}@illinois.edu} \\
{\small $^{2}$IBM T.J. Watson Research Center, NY, USA}\\
	\email{ \{Honghui.Shi, rsferis, jinjun\}@us.ibm.com}
}

\maketitle

\section{Differences with Related Works}
\subsubsection{RCNN}
There are two major differences between our DCR module and RCNN. First, our DCR module is an end-to-end classifier. We use softmax classifier on top of the CNN feature where as RCNN trains another SVM using CNN features. Second, the motivation is different. The purpose of RCNN is to classify each region, but the purpose of DCR module is to correct false positives produced by base detectors. The difference in motivation results in different sampling heuristic. RCNN samples a large batch of foreground and background with some fixed ratio to achieve a good balance for training classifier. Our DCR module not only samples foreground and background, but also pay attention to samples that Faster RCNN makes ``ridiculous'' mistakes (hard false positives).

\subsubsection{Hard Example Selection in Deep Learning}
Hard example mining is originally used for optimizing SVMs to achieve the global optimum. In \cite{shrivastava2016training}, an Online Hard Example Mining (OHEM) algorithm is proposed to train Fast RCNN. Instead of sampling the minibatch randomly, \cite{shrivastava2016training} samples ROIs that have the top losses (sum of classification and localization loss) with respect to the current set of parameters. \cite{liu2016ssd} uses a similar online approach, but instead of using hard examples with largest losses, \cite{liu2016ssd} further imposes a restriction on the ratio of foreground and background in hard examples. The main difference is that hard examples may not always be hard false positives. DCR module focuses all its attention to deal with hard false positive which means it is more task-specific than hard example selection methods. Another difference between OHEM and our approach is that OHEM take both classification and localization loss into account whereas DCR module only considers classification.

\subsubsection{Focal Loss (FL)}
FL \cite{lin2017focal} is designed to down-weight the loss of well-classified examples by adding an exponential term related to the probability of ground truth class, \ie $\text{FL}(p_t)=-(1-p_t)^{\gamma}\text{log}(p_t)$, where $\gamma$ is a tunable parameter specifying how much to down-weight. The motivation of FL is to use a dense set of predefined anchors on all possible image locations without region proposals as well as the sampling step. Since background dominants in this large set of boxes, FL ends up down-weighting most of losses for backgrounds instead of focusing on hard false positives.

\section{More Discussions}

Our DCR module demonstrates extremely good performance in suppressing false positives. Fig \ref{fig:motivation} (a) compares total number of false positives on the VOC2007 test set. With our DCR module, the number of hard false is reduced by almost three times (orange). 
The inference time of DCR module is proportional to the number of proposals, input size and network depth. Table \ref{ablation} (d), (e) compare the running time and the best model (DCR with depth 152) runs slower than the baseline Faster RCNN at the speed of 1.39 s/image on 1080 Ti GPU. However, this paper focuses more on the analysis of failure case of object detectors and accuracy boost, improvement to speed will be studied in the future.

\subsubsection{Error Analysis}
\begin{figure}
	\centering
	\includegraphics[width=1\textwidth]{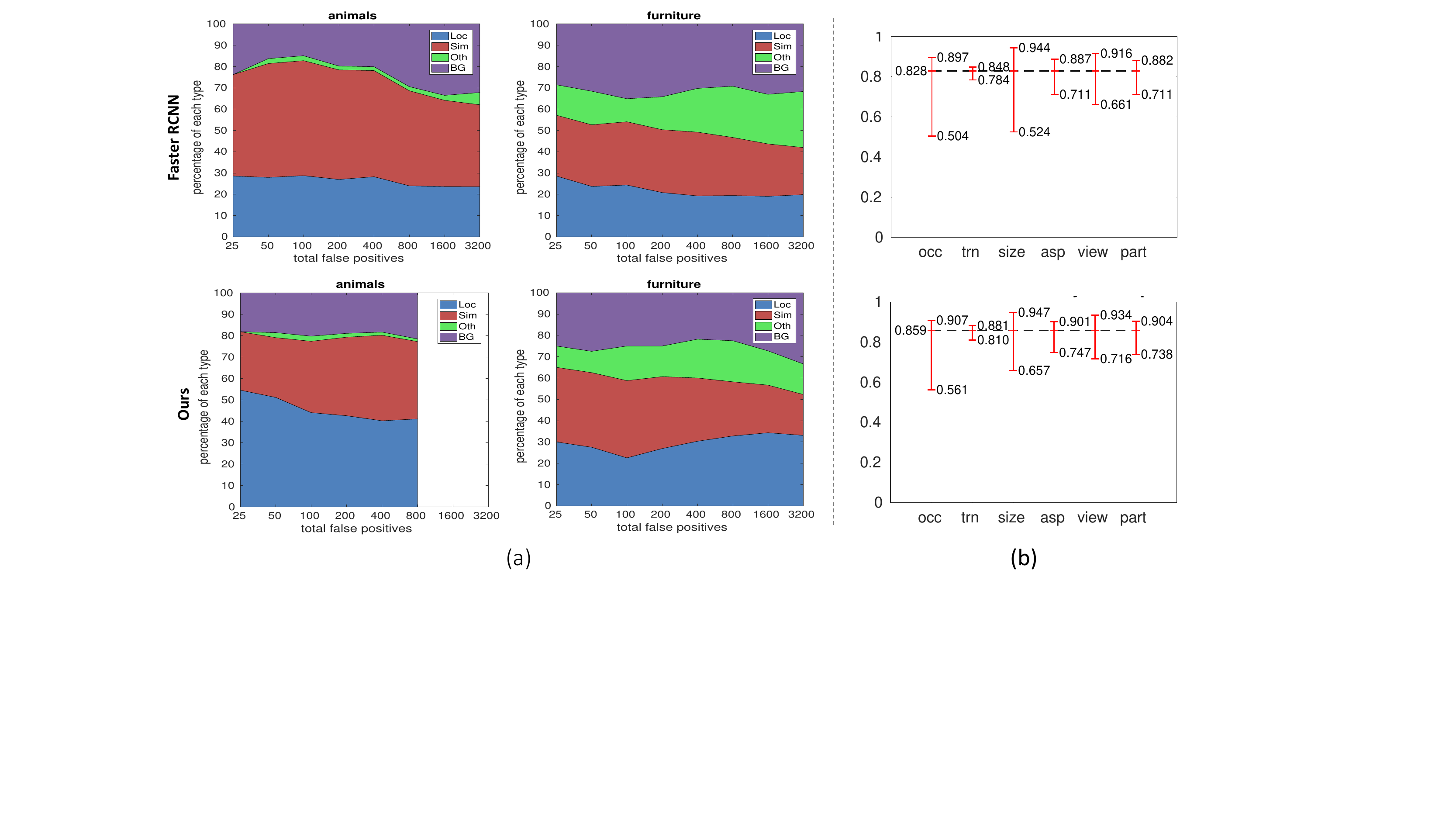}
	\caption{Analysis results between Faster RCNN (top row) and our methods (bottom row) by \cite{hoiem2012diagnosing}. Left of the dashed line: distribution of top false positive types. Right of the dashed line: sensitivity to object characteristics.}
	\label{fig:analysis}
\end{figure}

Following \cite{girshick2014rich}, we also use the detection analysis tool from \cite{hoiem2012diagnosing}, in order to gather more information of the error mode of Faster RCNN and DCR module. Analysis results are shown in Fig \ref{fig:analysis}. 

Fig \ref{fig:analysis} (a) shows the distribution of top false positive types as scores decrease. False positives are classified into four categories: (1) Loc: IOU with ground truth boxes is in the range of $[0.1, 0.5)$; (2) Sim: detections have at least 0.1 IOU with objects in predefined similar classes, \eg dog and cat are similar classes; (3) Oth: detections have at least 0.1 IOU with objects not in predefined similar classes; (4) BG: all other false positives are considered background. We observe that comparing with Faster RCNN, DCR module has much larger ratio of localization error and the number of false positives is greatly reduced on some classes, \eg in the animal class, the number of false positives is largely reduced by 4 times and initial percentage of localization error increases from less than 30\% to over 50\%. This statistics are consistent with motivations to reducing classification errors by reducing number of false positives. 

Fig \ref{fig:analysis} (b) compares the sensitivity of Faster RCNN and DCR to object characteristics. \cite{hoiem2012diagnosing} defines object with six characteristics: (1) occ: occlusion, where an object is occluded by another surface; (2) trn: truncation, where there is only part of an object; (3) size: the size of an object measure by the pixel area; (4) asp: aspect ratio of an object; (5) view: whether each side of an object is visible; (6) part: whether each part of an object is visible. Normalized AP is used to measure detectors performance and more details can be found in \cite{hoiem2012diagnosing}. In general, the higher the normalized AP, the better the performance. The difference between max and min value measure the sensibility of a detector, the smaller the difference, the less sensible of a detector. We observe that DCR improves normalized AP and sensitivity on all types of object and improves sensitivity significantly on occlusion and size. This increase came from the adaptive field of DCR, since DCR can focus only on the object area, making it less sensible to occlusion and size of objects.

\section{Visualization}
\begin{figure}
	\centering
	\includegraphics[width=0.95\textwidth]{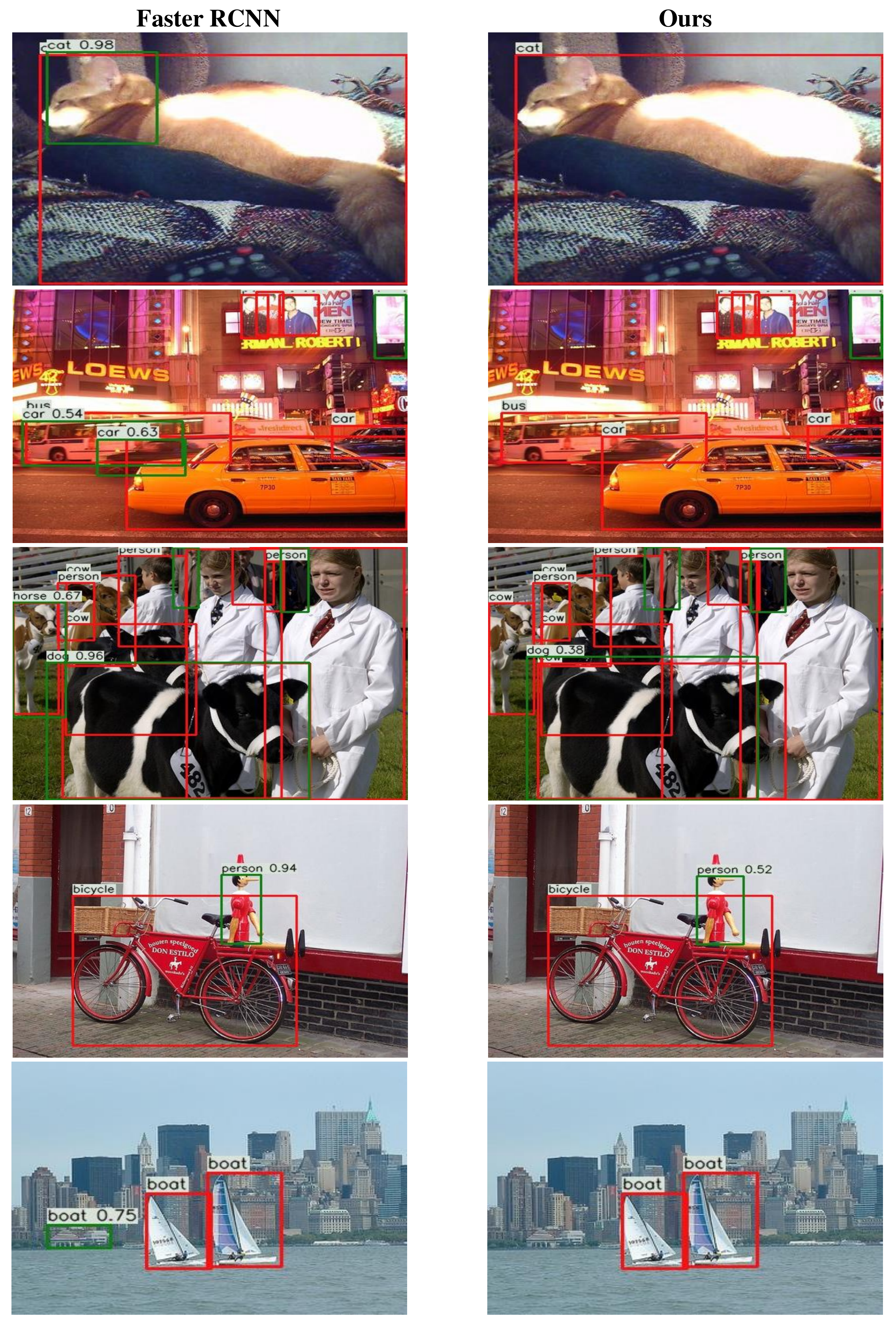}
	\caption{Comparison of hard false positives with confidence score higher than 0.3 between Faster RCNN and Our methods. Red box: ground truth object. Green box: hard false positive.}
	\label{fig:visualize_fp}
\end{figure}

We visualize all false positives with confidence larger than 0.3 for both Faster RCNN and our DCR module in Fig \ref{fig:visualize_fp}. We observe that the DCR module successfully suppresses all three kinds of hard false positives to some extends. 

The first image shows reducing the first type of false positives (part of objects). Faster RCNN (left) classifies the head of the cat with a extremely high confidence (0.98) but it is eliminated by the DCR module. 

The second to the fourth images demonstrate situations of second type of false positives (similar objects) where most of false positives are suppressed (``car'' in the second image and ``horse'' in the third image). However, we find there still exists some limitations, \eg the ``dog'' in the third image where it is supposed to be a cow and the ``person'' in the fourth image. Although they are not suppressed, their scores are reduced significantly (0.96 $\rightarrow$ 0.38 and 0.92 $\rightarrow$ 0.52 respectively) which will also improve the overall performance. It still remains an open questions to solve these problems. We hypothesize that by using more training data of such hard false positives (\eg use data augmentation to generate such samples).

The last image shows example for the third type of false positives (backgrounds). A part of background near the ground truth is classified as a ``boat'' by the Faster RCNN and it is successfully eliminated by our DCR module.

\bibliographystyle{splncs}
\bibliography{egbib}
\end{document}